\documentclass[conference]{IEEEtran}
\IEEEoverridecommandlockouts
\usepackage{bm}
\usepackage{cite}
\usepackage{svg}
\usepackage{amsmath,amssymb,amsfonts}
\usepackage{graphicx}
\usepackage{textcomp}
\usepackage{xcolor}
\usepackage{caption}
\usepackage{subcaption}
\usepackage{algorithm}
\usepackage{algorithmic}
\usepackage{hyperref}
\def\BibTeX{{\rm B\kern-.05em{\sc i\kern-.025em b}\kern-.08em
    T\kern-.1667em\lower.7ex\hbox{E}\kern-.125emX}}

\newlength{\myfigwidth}
\ifCLASSOPTIONonecolumn
	\AtBeginDocument{\setlength{\myfigwidth}{0.8\linewidth}}
\else
	\AtBeginDocument{\setlength{\myfigwidth}{\linewidth}}
\fi
\newcommand{\myunit}[1]{%
	\ifmmode
		\mathrm{#1}
	\else
		$ \mathrm{#1} $% <-this % stops a space
	\fi}
\newcounter{MYalgorithmic}

{%
	\vspace*{3pt}
	\hrule height 1pt}
\newcommand{\MYlabel}[1]{\def\@currentlabel{\theALG@line}\label{#1}}

\definecolor{myblue}{rgb}{0,0.4980,1} % Azure
\definecolor{myred}{rgb}{0.8706,0.1608,0.0627} % Chinese red

\begin{document}

\title{SemDynReg: Semantics-Guided Deformation Regularization for Dynamic 3D Gaussian Splatting\\
}
% Original author/affiliation block kept for reference:
% \author{
% \author{\IEEEauthorblockN{Ruitao Chen\IEEEauthorrefmark{1},
% Mozhang Guo\IEEEauthorrefmark{1}, and Jinge Li\IEEEauthorrefmark{2}}
% \IEEEauthorblockA{\IEEEauthorrefmark{1}Department of Electrical and Computer Engineering,
% Western University, London, Canada\\
% \IEEEauthorrefmark{2}Department of Computing and Software, McMaster University, Hamilton, Canada\\
% E-mail: rchen328@uwo.ca, mguo224@uwo.ca, and lij269@mcmaster.ca}}
% }

\author{
\author{\IEEEauthorblockN{Ruitao Chen, Mozhang Guo, and Jinge Li}
\IEEEauthorblockA{}}
}

% \author{
% Ruitao Chen,
% Mozhang Guo, and Jinge Li \\
% E-mail: \{rchen328, mguo224\}@uwo.ca; lij269@mcmaster.ca
% }

\maketitle

\begin{abstract}
Deformable 3D Gaussian Splatting (3DGS) has emerged as an efficient approach for rendering dynamic scenes in a wide range of 3D applications. 
However, existing deformation field–based approaches largely lack explicit object-level modeling, 
often resulting in inconsistent Gaussian deformations within individual objects and unwanted coupling between different objects.
To address this limitation, we introduce a semantics-guided framework that enforces dynamic regularization at the object level, 
aiming to achieve spatially consistent object-wise deformation. 
Specifically, we first extract segmentation masks using the Segment Anything Model (SAM) and derive semantic features from input images. 
An object-ID map is then constructed via feature relevance matching with a predefined object dictionary. 
Guided by this object-ID map, we identify the pixel-wise top-$k$ contributing Gaussians for each object and impose consistency regularization on their deformation parameters, 
including position, scale, and rotation. 
Unlike prior methods that learn deformation fields without explicit object-level constraints, our approach incorporates semantic cues to guide deformation behavior at the object level. 
Experimental results demonstrate that our semantics-aware regularization improves object-level deformation consistency and outperforms baseline methods in rendering quality, achieving higher PSNR and SSIM and lower LPIPS in dynamic 3DGS rendering.
Our project page is available at \url{https://dyn-reg-3dgs.github.io/}.
\end{abstract}

% \begin{IEEEkeywords}
% Dynamic 3D Gaussian Splatting, semantics-guided regularization, object-level deformation consistency, Segment Anything Model (SAM), CLIP, dynamic scene rendering
% \end{IEEEkeywords}

\section{Introduction}
Dynamic 3D Gaussian Splatting (3DGS) has recently emerged as an efficient representation for modeling dynamic scenes, 
attracting significant attention in computer vision and graphics due to its role in applications 
such as autonomous driving simulation, virtual reality, and 360-degree video generation \cite{wu20244d,yang2024deformable,zhou2024drivinggaussian,hess2025splatad}.
Such scenes often involve dynamic objects, such as vehicles and humans, that exhibit structured and complex motion patterns. 
Accurately modeling the motion and deformation of these dynamic objects is essential for producing realistic and immersive visual experiences. 
For example, in a driving scene, realistic rendering of moving vehicles requires capturing their motion trajectories and appearance variations across viewpoints \cite{zhou2024drivinggaussian,hess2025splatad}, while modeling pedestrians involves representing articulated motions such as walking and turning \cite{yu2026Part}.

To model dynamics in 3DGS, deformable approaches represent scene motion by learning continuous deformation fields over Gaussians \cite{yang2024deformable,wu20244d,lu20243d}. 
Despite its effectiveness, existing deformation field–based approaches typically operate in an object-agnostic manner and do not explicitly account for object-level structure. 
As a result, the learned deformations may vary inconsistently across Gaussians within the same object, as well as across different objects. 
This often leads to visual artifacts and reduced realism, such as distortions in the structure of moving vehicles or incoherent deformation of articulated body parts. 
Moreover, for applications that require understanding motion patterns, 
the lack of structured constraints on deformation makes it difficult to capture reliable and coherent motion trajectories.

To address these issues, we propose a semantics-guided approach that introduces object-level regularization into dynamic 3DGS.
Instead of treating Gaussians independently, our approach leverages semantic cues to group Gaussians belonging to the same object and encourages them to deform coherently over time.
Specifically, we extract segmentation masks using the Segment Anything Model (SAM) \cite{kirillov2023segment} and encode their semantics using the Contrastive Language-Image Pre-Training (CLIP) model \cite{radford2021learning} to form a feature map, 
which is matched with a predefined object dictionary to construct an object-ID map.
Leveraging this object-level association, we select the pixel-wise top-$k$ contributing Gaussians per object and impose consistency constraints on their deformation parameters.
This object-centric regularization mitigates the inconsistencies of purely deformation field–based methods, yielding more structured and physically plausible motion, while remaining integrable into existing dynamic 3DGS pipelines.
In practice, the required semantic features are often already available in semantic-aware applications such as object editing, segmentation, scene understanding, 
and language-guided interaction \cite{zhou2024feature}, allowing our approach to leverage these features with minimal additional cost.

The key contributions of this paper are summarized as follows:
\begin{itemize}
\item We introduce a semantics-guided regularization approach for dynamic 3DGS that enforces object-level consistency in deformation by
identifying and grouping object-associated Gaussians based on semantic cues and applying regularization to their deformation parameters.
\item Rather than relying on a single dominant Gaussian, we propose to select the pixel-wise top-$k$ contributing Gaussians for each object 
to ensure more comprehensive identification of object-associated Gaussians, which enhances the robustness of regularization and leads to improved deformation consistency.
\item We evaluate our method on dynamic driving scenes and show that the proposed semantics-aware regularization 
significantly improves object-level consistency in deformation, leading to enhanced visual quality and more realistic motion representation.
\end{itemize}

This paper is structured as follows. 
Section~\ref{sec:related_work} reviews related work on dynamic 3D Gaussian Splatting and 
semantic object-level scene modelling. 
Section~\ref{sec:method} presents in detail the proposed semantic-guided object-level regularization 
framework. 
Section~\ref{sec:experiments} describes the experimental setup and evaluates the effectiveness of the proposed method on dynamic driving scenes. 
Finally, Section~\ref{sec:conclusion} 
concludes the paper and discusses future directions.

\section{Related Work}\label{sec:related_work}
% \subsection{Dynamic 3D Gaussian Splatting and Regularization}
% Related work.
% \subsection{Object-level and Structured Scene Modeling}
% Related work.
\subsection{Dynamic 3D Gaussian Splatting}
3D Gaussian Splatting (3DGS) has emerged as an efficient explicit representation for real-time novel-view synthesis. By representing a scene as a set of anisotropic 3D Gaussians and rendering them through visibility-aware splatting, 3DGS achieves high-quality rendering with real-time performance \cite{kerbl20233d}. To handle dynamic scenes, recent methods extend 3DGS with time-dependent Gaussian motion, deformation fields, sparse motion control, and spacetime representations. Dynamic 3D Gaussians allow Gaussians to move and rotate over time while maintaining persistent color, opacity, and size \cite{luiten2024dynamic}. 4D Gaussian Splatting represents dynamic scenes using 3D Gaussians together with 4D neural voxel features, from which a lightweight decoder predicts time-dependent Gaussian deformations \cite{wu20244d}. Deformable 3D Gaussians reconstruct monocular dynamic scenes by optimizing Gaussians in a canonical space and learning deformation fields that transform them over time \cite{yang2024deformable}. Other methods further introduce geometry-aware deformation modeling, sparse control points with locally rigid motion constraints, and spacetime Gaussian representations to improve temporal coherence and rendering quality \cite{lu20243d, huang2024sc, li2024spacetime}. These works demonstrate the effectiveness of temporal deformation, motion regularization, and geometric or spacetime priors for dynamic 3DGS rendering.

\subsection{Semantic and Object-Aware Gaussian Representations}

Semantic 3DGS methods enhance Gaussian representations with language, instance-level, or object-level information. LangSplat distills CLIP-based language features into 3D Gaussian representations, using a scene-wise language autoencoder and SAM-derived hierarchical semantics to support efficient open-vocabulary querying in 3D scenes \cite{qin2024langsplat}. Gaussian Grouping augments 3D Gaussians with learnable identity encodings and supervises them using SAM-generated 2D masks together with 3D spatial consistency regularization, enabling open-world 3D segmentation and object-level scene editing \cite{ye2024gaussian}. Feature 3DGS augments 3D Gaussian representations with distilled feature fields from 2D foundation models, enabling novel-view semantic segmentation, promptable segmentation, and language-guided editing \cite{zhou2024feature}. OpenGaussian improves point-level open-vocabulary understanding by learning 3D-consistent Gaussian instance features from SAM masks, discretizing them with a two-stage codebook, and associating 3D instances with CLIP semantics through 3D--2D feature association \cite{wu2024opengaussian}. More recent methods strengthen object-aware Gaussian modeling through object anchors, object-level codebooks, and Gaussian-level semantic features for scene reconstruction and instance segmentation \cite{zhu2025objectgs, zhu2025rethinking}. These works show that semantic and object-level information can be effectively incorporated into Gaussian representations.

\subsection{Semantic Guidance for Dynamic Gaussian Modeling}

Recent studies have started to connect semantic or object-level understanding with dynamic Gaussian representations. 4D LangSplat extends language Gaussian fields to dynamic scenes by using MLLM-generated object-wise video captions and language embeddings to learn 4D language fields, enabling time-agnostic and time-sensitive open-vocabulary queries \cite{li20254d}. CoDa-4DGS incorporates context and temporal deformation awareness for autonomous-driving scenes by self-supervising Gaussian semantic features and aggregating contextual and deformation cues for deformation compensation \cite{song2025coda}. Object-aware dynamic methods such as AD-GS introduce learnable motion models based on locality-aware B-spline curves and global-aware temporal functions for self-supervised driving-scene rendering, decomposing scenes into objects and background and representing dynamic objects with Gaussians \cite{xu2025ad}. Part-aware Gaussian methods further show that object or part structure can provide useful priors for modeling articulated motion and preserving physically plausible deformation \cite{yu2026Part}.

In general, existing dynamic 3DGS methods mainly improve deformation modeling through temporal, geometric, motion, or control-based priors, while semantic 3DGS methods mainly focus on scene understanding, open-vocabulary querying, segmentation, or editing. Recent semantic-aware and object-aware dynamic methods begin to bridge these directions, but explicit semantic object-level regularization of Gaussian deformation parameters remains less explored in general dynamic 3DGS rendering. In this work, we use semantic object membership to guide deformation regularization, encouraging Gaussians associated with the same object to deform coherently in position, scale, and rotation.

\section{Method}\label{sec:method}
\begin{figure*}[t]
	\centering
	\includegraphics[width=\linewidth]{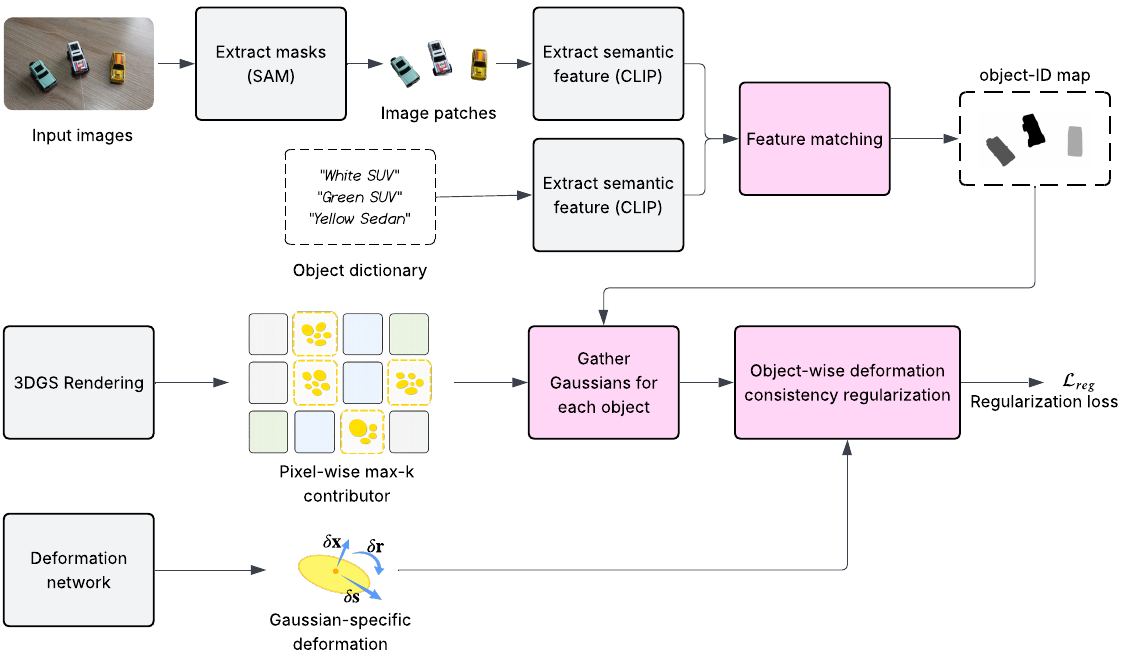}
	\caption{Overview of the proposed SemDynReg pipeline.}
	\label{fig:overview_pipeline}
\end{figure*}

Our method introduces a semantics-guided approach to enforce object-level consistency in dynamic 3D Gaussian Splatting. By leveraging semantic cues, we establish object-level associations among Gaussians and impose consistency constraints on their deformation to achieve coherent motion. The overall framework integrates semantic feature extraction, object identification, Gaussian association, and deformation regularization into a unified pipeline, as illustrated in Fig.~\ref{fig:overview_pipeline}.

\subsection{Semantic Feature Extraction with SAM and CLIP}\label{sec:semantic_feature_extraction}

As illustrated in the \emph{overview of our pipeline} (Fig.~\ref{fig:overview_pipeline}), the first stage extracts semantic features from input images to enable object-level reasoning. This stage consists of two steps: object mask extraction using SAM and semantic feature encoding using CLIP.

Given an input image \( I \in \mathbb{R}^{H \times W \times 3} \), we first apply SAM \cite{kirillov2023segment} to obtain a set of candidate segmentation masks. Following prior works \cite{kirillov2023segment}, we remove redundant masks based on their predicted Intersection-over-Union (IoU) scores, stability scores, and pairwise overlap. This filtering process produces a set of non-overlapping, high-quality object-level masks:
\begin{equation}
\mathcal{M} = \{ M_j \mid M_j \in \{0,1\}^{H \times W} \}_{j=1}^{K}.
\end{equation}

For each mask \( M_j \), we extract the corresponding image region and encode its semantic representation using CLIP \cite{radford2021learning}:
\begin{equation}
\mathbf{f}_j = \mathrm{CLIP}(I \odot M_j), \quad \mathbf{f}_j \in \mathbb{R}^{512},
\end{equation}
where \( \odot \) denotes element-wise masking.

Since the masks are non-overlapping, each pixel is uniquely associated with a single mask and directly inherits its corresponding semantic feature. This results in a dense semantic feature map \( \mathbf{F} \in \mathbb{R}^{H \times W \times 512} \), where each spatial location is represented by a 512-dimensional CLIP feature.

This feature map provides semantic guidance for subsequent object identification and Gaussian association, forming the basis for object-level deformation regularization.
\subsection{Object-ID Map Construction with Object Dictionary}\label{sec:object_id_map_construction}

Given the semantic feature map \( \mathbf{F} \in \mathbb{R}^{H \times W \times 512} \) from Sec.~3.1, we construct an object-ID map by matching pixel-wise features to a predefined object dictionary.

We define an object dictionary as a set of textual prompts representing objects of interest:
\begin{equation}
\mathcal{D} = \{ t_i \}_{i=1}^{C},
\end{equation}
where each prompt \( t_i \) (e.g., \emph{``white SUV''}, \emph{``green SUV''}) is encoded into a semantic embedding
\(
\mathbf{e}_i = \mathrm{CLIP}_{\text{text}}(t_i) \in \mathbb{R}^{512}.
\)

For a pixel \( v \) (spatial location in the image), we compute its relevance to each object by comparing its feature \( \mathbf{F}(v) \) with the object embeddings. Following prior works\cite{qin2024langsplat, kerr2023lerf}, we introduce a set of generic context prompts \( \mathcal{C} = \{ \text{object}, \text{stuff}, \text{thing}, \text{texture} \} \) to compute a normalized relevance score:
\begin{equation}
r_i(v) = \min_{c \in \mathcal{C}} 
\frac{\exp\left( \mathbf{F}(v) \cdot \mathbf{e}_i \right)}
{\exp\left( \mathbf{F}(v) \cdot \mathbf{e}_i \right) + \exp\left( \mathbf{F}(v) \cdot \mathbf{e}_i^{c} \right)},
\end{equation}
where \( \mathbf{e}_i^{c} \) denotes the embedding of prompt \( t_i \) composed with context \( c \).

We then assign an object label to each pixel based on a threshold \( \tau \):
\begin{equation}
\mathrm{ID}(v) =
\begin{cases}
\arg\max_i \, r_i(v), & \text{if } \max_i r_i(v) > \tau, \\
C, & \text{otherwise},
\end{cases}
\end{equation}
where \(C\) is an irrelevant or background label, while objects of interest are indexed from 0 to \(C-1\).

By introducing a threshold on the relevance score, the method can more reliably distinguish objects of interest from irrelevant regions, leading to more accurate object identification.
This results in an object-ID map \( \mathrm{ID} \in \mathbb{R}^{H \times W} \), where each pixel is assigned either an object label or an irrelevant label.
\subsection{Top-$k$ Gaussian Selection and Object-level Grouping}\label{sec:topk_gaussian_selection}

Given the object-ID map \( \mathrm{ID} \in \mathbb{R}^{H \times W} \) from Sec.~\ref{sec:object_id_map_construction}, we associate Gaussians with objects by identifying the most influential Gaussians contributing to each pixel and grouping them accordingly.

Let \( \mathcal{N} \) denote the global set of all Gaussians in the scene. In 3D Gaussian Splatting\cite{kerbl20233d}, the color of a pixel \( \mathbf{p} \) is computed using alpha blending over Gaussians that overlap the pixel and are sorted in front-to-back depth order:
\begin{equation}
C(\mathbf{p}) = \sum_{i \in \mathcal{N}} T_i(\mathbf{p}) \, \alpha_i(\mathbf{p}) \, \mathbf{c}_i,
\end{equation}
where only Gaussians contributing to \( \mathbf{p} \) have non-zero terms, \( \alpha_i(\mathbf{p}) \) is the opacity of Gaussian \(i\) at pixel \( \mathbf{p} \), and \( T_i(\mathbf{p}) = \prod_{j<i} (1 - \alpha_j(\mathbf{p})) \) is the accumulated transmittance.

We define the contribution weight of Gaussian \(i\) to pixel \( \mathbf{p} \) as
\begin{equation}
w_i(\mathbf{p}) = T_i(\mathbf{p}) \, \alpha_i(\mathbf{p}).
\end{equation}

For each pixel, we select the indices of the \(k\) Gaussians with the largest contribution weights:
\begin{equation}
\mathcal{K}(\mathbf{p}) = \operatorname{TopK}_{i \in \mathcal{N}} \, w_i(\mathbf{p}), \quad |\mathcal{K}(\mathbf{p})| = k,
\end{equation}
where \( \mathcal{K}(\mathbf{p}) \) contains indices in the global Gaussian set \( \mathcal{N} \). This produces a dense top-$k$ Gaussian map of size \( H \times W \times k \).

To obtain object-level Gaussian groups, we leverage the object-ID map. For each object \( o \in \{0, \dots, C-1\} \), we collect the top-$k$ Gaussians from all pixels assigned to that object:
\begin{equation}
\mathcal{G}_o = \{ \mathcal{K}(\mathbf{p}) \mid \mathrm{ID}(\mathbf{p}) = o \}.
\end{equation}

Since the selected top-$k$ Gaussians dominate the alpha blending process for each pixel, aggregating them over pixels belonging to the same object establishes a correspondence between 2D object regions and their associated 3D Gaussians.
\subsection{Deformation Regularization with Object-level Constraints} \label{sec:deformation_regularization}

Building on the object-level Gaussian grouping in Sec.~\ref{sec:topk_gaussian_selection}, we introduce a regularization term that enforces consistent deformation among Gaussians belonging to the same object.

We adopt the deformation model from deformable 3D Gaussian Splatting \cite{yang2024deformable}, where each Gaussian \( i \in \mathcal{N} \) is associated with three deformation parameters: translation \( \delta \mathbf{x}_i \in \mathbb{R}^3 \), rotation \( \delta \mathbf{r}_i \in \mathbb{R}^3 \), and scaling \( \delta \mathbf{s}_i \in \mathbb{R}^3 \), as shown in Fig.~1. These parameters model the dynamic transformation of each Gaussian. The proposed regularization is general and can be extended to other deformation parameterizations.

For each object \( o \in \{0, \dots, C-1\} \), we consider the set of associated Gaussians \( \mathcal{G}_o \) obtained in Sec.~\ref{sec:topk_gaussian_selection}. Let \( \bar{\delta \mathbf{x}}_o \), \( \bar{\delta \mathbf{r}}_o \), and \( \bar{\delta \mathbf{s}}_o \) denote the mean deformation parameters over all Gaussians in \( \mathcal{G}_o \). We enforce consistency by penalizing deviations from the object-level mean using mean squared error:
\begin{equation}
\begin{aligned}
\mathcal{L}_{\text{reg}}^{(o)} =
& \ \lambda_x \frac{1}{|\mathcal{G}_o|} \sum_{i \in \mathcal{G}_o} 
\| \delta \mathbf{x}_i - \bar{\delta \mathbf{x}}_o \|_2^2 \\
& + \lambda_r \frac{1}{|\mathcal{G}_o|} \sum_{i \in \mathcal{G}_o} 
\| \delta \mathbf{r}_i - \bar{\delta \mathbf{r}}_o \|_2^2 \\
& + \lambda_s \frac{1}{|\mathcal{G}_o|} \sum_{i \in \mathcal{G}_o} 
\| \delta \mathbf{s}_i - \bar{\delta \mathbf{s}}_o \|_2^2.
\end{aligned}
\end{equation}

To avoid over-constraining objects with negligible motion and to reduce computational overhead, we optionally exclude static objects from regularization. An object is considered static if the magnitude of its mean deformation is below a threshold, e.g., \( \| \bar{\delta \mathbf{x}}_o \|_2 \leq \tau_d \), in which case the regularization for that object is skipped.

The overall regularization loss is obtained by summing over all objects:
\begin{equation}
\mathcal{L}_{\text{reg}} = \sum_{o} \mathcal{L}_{\text{reg}}^{(o)}.
\end{equation}

This loss is combined with the original RGB reconstruction loss of 3DGS:
\begin{equation}
\mathcal{L} = \mathcal{L}_{\text{rgb}} + \lambda \mathcal{L}_{\text{reg}},
\end{equation}
where \( \lambda \) controls the strength of the regularization. In practice, the regularization term is activated after an initial training phase (e.g., after 10k iterations) to allow the deformation field to stabilize before enforcing object-level constraints.

We note that the proposed regularization is primarily designed for objects with approximately rigid motion, such as vehicles, where enforcing consistent deformation across associated Gaussians is well aligned with the underlying structure. For non-rigid objects (e.g., humans), a single object-level constraint may be overly restrictive due to articulated motion. However, the approach can be extended by decomposing objects into semantically meaningful components (e.g., body parts) and applying the regularization within each component, enabling adaptation to more complex non-rigid scenarios.

\section{Experiment}\label{sec:experiments}

\subsection{Experiment Setup}
We evaluate the proposed SemDynReg within a monocular dynamic scene reconstruction setting based on deformable 3D Gaussian Splatting.

\textbf{Dataset.}
We construct a private dataset consisting of controlled dynamic scenes, as public datasets typically do not provide the object-level semantic structure required for evaluating semantics-guided deformation consistency. Each scene contains three miniature toy vehicles placed on a planar surface, where one vehicle undergoes motion while the others remain static. The sequences are captured using a moving monocular camera over multiple time steps, introducing both object motion and viewpoint changes. This setup enables clear analysis of deformation consistency for dynamic versus static objects under realistic viewing conditions.

\textbf{Implementation Details.}
Our implementation follows the standard setup of deformable 3D Gaussian Splatting and adopts semantic feature extraction settings consistent with prior language-guided 3DGS works. For semantic processing, we use OpenCLIP with a ViT-B/16 backbone to extract image and text embeddings, and SAM with a ViT-H backbone to generate 2D segmentation masks.

Training is performed for 40,000 iterations. We use an initial warm-up phase of 3,000 iterations where only the canonical 3D Gaussians are optimized without deformation. From 3k to 10k iterations, the deformation field is enabled but trained without regularization. The proposed regularization term is activated from 10k to 40k iterations and jointly optimized with the reconstruction loss.

All experiments are conducted at a resolution of 1440×1080. Training a scene with the proposed regularization takes approximately 25 minutes on a single NVIDIA RTX 3090 GPU and requires around 4 GB of GPU memory.

\textbf{Evaluation Strategy.}
We focus on qualitative evaluation to assess the effectiveness of the proposed object-level deformation regularization. For each scene, we compare rendering results between the baseline (without regularization) and our method, including ground truth frames, object-ID maps, rendered outputs, and zoomed-in views of the dynamic object across time.

Improvements are assessed based on reduced deformation artifacts, sharper object appearance, and more coherent motion over time.

\subsection{Result Discussion}
\subsubsection{Results on Dynamic Scene 1}

\begin{figure*}[t]
	\centering
	\includegraphics[width=\linewidth]{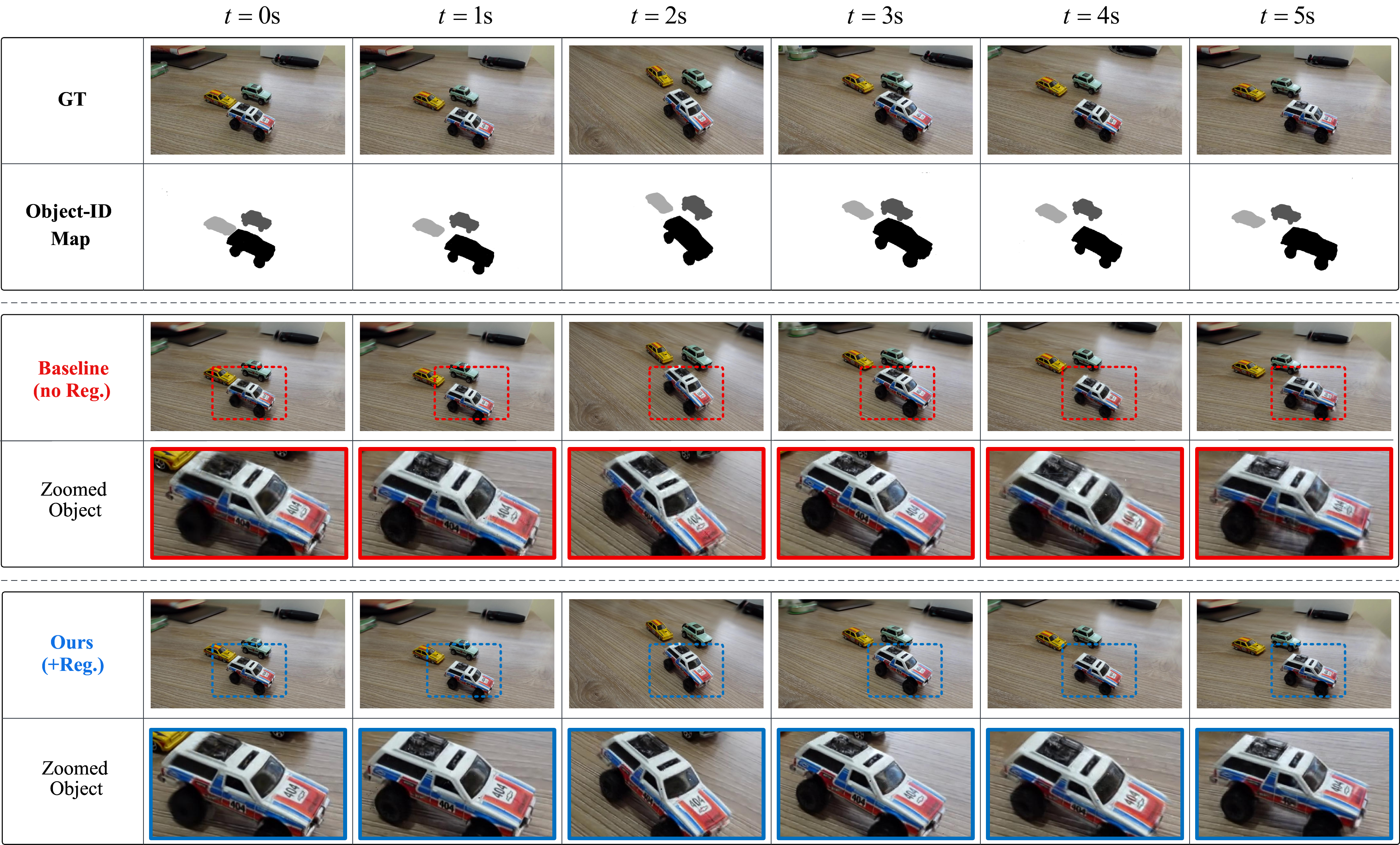}
	\caption{Results on dynamic scene 1 (white SUV as the moving object).}
	\label{fig:result_discussion_0101}
\end{figure*}

Fig.~\ref{fig:result_discussion_0101} presents the results on the first dynamic scene, 
where the white SUV is the primary moving object. The object-ID maps show that the proposed semantic matching step can consistently 
identify the vehicles across different time steps, providing reliable object-level associations for deformation regularization.
Compared with the baseline without regularization, our method produces a noticeably sharper and more stable reconstruction of 
the moving white SUV. The improvement is particularly evident at $t = 0\text{s}$ and $t = 5\text{s}$. 
In the baseline results, the SUV exhibits visible blur around the body and wheels, and the boundary of the vehicle appears less 
well-defined. Fine structures such as the window regions and printed patterns on the car body are partially smeared.

In contrast, our method preserves clearer object boundaries and finer structural details. 
At $t = 0\text{s}$, the edges of the SUV and the contrast between different colored regions on the car body are more distinct. 
At $t = 5\text{s}$, the improvement remains consistent, with better preservation of the vehicle shape and reduced motion-induced artifacts.
 These visual observations are further supported by the quantitative results in Table~\ref{tab:metrics_zoom_1}. 
 Averaged across all time steps, the proposed method improves SSIM from 0.841 to 0.888 and PSNR from 26.459\,dB to 29.615\,dB, 
 while reducing LPIPS from 0.141 to 0.109. The zoomed-in views further highlight that the deformation of the SUV is more coherent over time, leading to improved visual quality.

\begin{table}[t]
\centering
\caption{Quantitative comparison on the zoomed object across time in scene 1.}
\label{tab:metrics_zoom_1}
\small
\setlength{\tabcolsep}{3pt}
\begin{tabular}{c|cccccc|c}
\hline
 & $t=0$s & $t=1$s & $t=2$s & $t=3$s & $t=4$s & $t=5$s & Avg. \\
\hline

\multicolumn{8}{c}{\textbf{SSIM $\uparrow$}} \\
\hline
Baseline & 0.829 & 0.861 & 0.831 & 0.889 & 0.818 & 0.819 & 0.841 \\
Ours     & 0.905 & 0.907 & 0.880 & 0.908 & 0.857 & 0.874 & \textbf{0.888} \\
\hline

\multicolumn{8}{c}{\textbf{PSNR $\uparrow$}} \\
\hline
Baseline & 24.201 & 27.460 & 26.220 & 30.169 & 25.273 & 25.431 & 26.459 \\
Ours     & 29.190 & 31.767 & 29.417 & 30.890 & 27.654 & 28.772 & \textbf{29.615} \\
\hline

\multicolumn{8}{c}{\textbf{LPIPS $\downarrow$}} \\
\hline
Baseline & 0.139 & 0.112 & 0.154 & 0.106 & 0.169 & 0.168 & 0.141 \\
Ours     & 0.088 & 0.082 & 0.124 & 0.098 & 0.138 & 0.124 & \textbf{0.109} \\
\hline

\end{tabular}
\end{table}

\subsubsection{Results on Dynamic Scene 2}

\begin{figure*}[t]
	\centering
	\includegraphics[width=\linewidth]{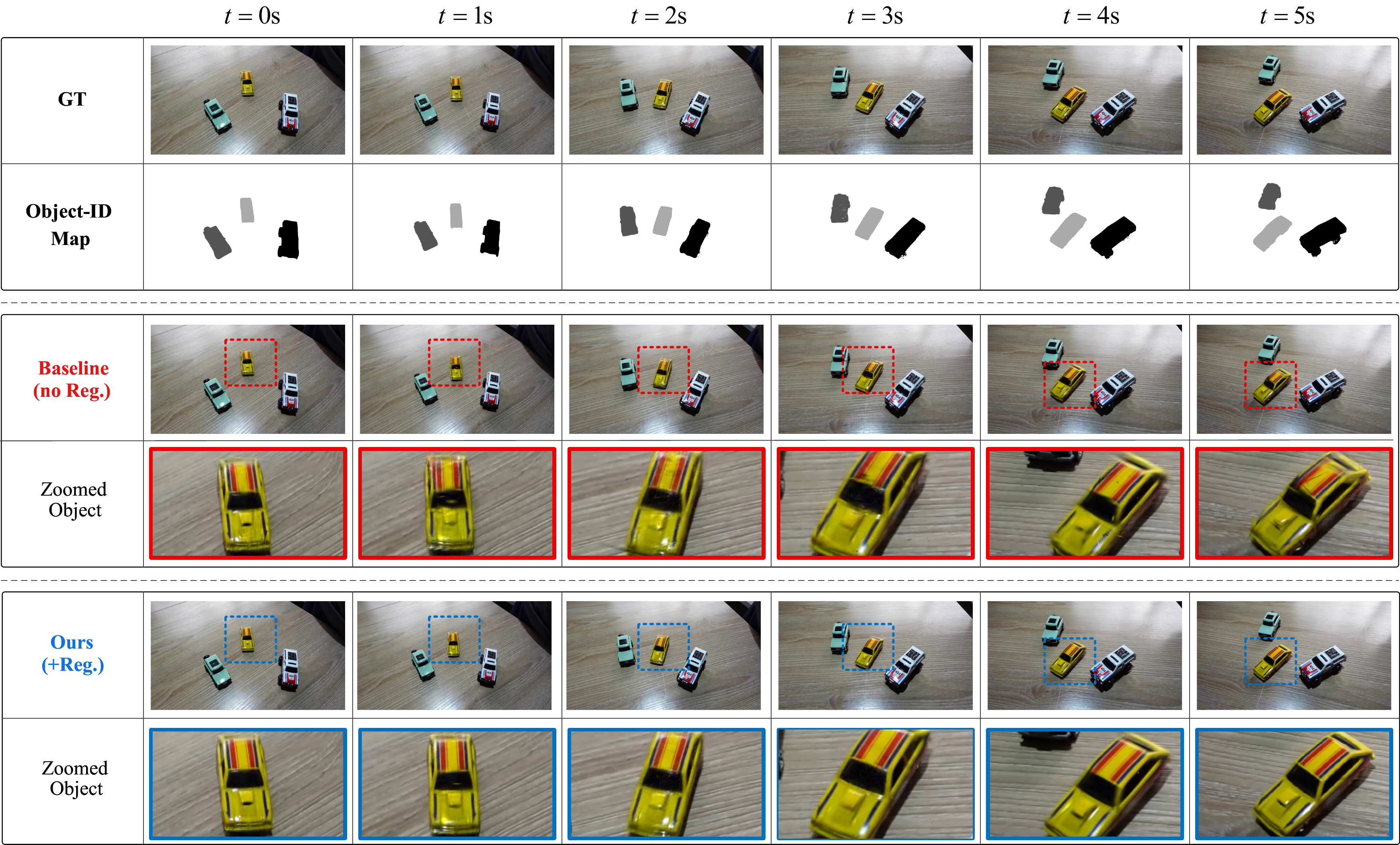}
	\caption{Results on dynamic scene 2 (yellow sedan as the moving object).}
	\label{fig:result_discussion_dynamic}
\end{figure*}

Fig.~\ref{fig:result_discussion_dynamic}shows the results on the second dynamic scene, where the yellow sedan is the moving object. 
The object-ID maps again demonstrate reliable identification of individual vehicles, enabling effective object-level grouping of Gaussians.
Across all time steps, the proposed method consistently improves the rendering quality of the moving yellow sedan compared to the baseline.
In the baseline results, the vehicle exhibits noticeable blur and loss of detail, particularly on the roof and along object boundaries. For example, at $t = 2\text{s}$ and $t = 3\text{s}$, the red stripe on the roof appears less stable and slightly smeared, and the edges of the vehicle are less clearly defined.

With the proposed regularization, the rendered sedan shows sharper edges and improved detail fidelity across all frames. The roof stripe is more clearly preserved, and the overall vehicle structure appears more stable under motion and viewpoint changes. These visual observations are further supported by the quantitative results in Table~\ref{tab:metrics_zoom_2}. Averaged across all time steps, the proposed method improves SSIM from 0.774 to 0.861 and PSNR from 23.070\,dB to 26.905\,dB, while reducing LPIPS from 0.151 to 0.099. This indicates that the deformation of Gaussians associated with the object is more consistent over time, resulting in improved reconstruction quality.

Across both scenes, the proposed semantics-guided regularization improves the deformation consistency of dynamic objects, leading to reduced motion blur, better preservation of fine structures, and more stable object appearance over time. At the same time, the visual quality of static objects remains largely unchanged, indicating that the method introduces minimal side effects while selectively enhancing dynamic regions.

\begin{table}[t]
\centering
\caption{Quantitative comparison on the zoomed object across time in scene 2.}
\label{tab:metrics_zoom_2}
\small
\setlength{\tabcolsep}{3pt}
\begin{tabular}{c|cccccc|c}
\hline
 & $t=0$s & $t=1$s & $t=2$s & $t=3$s & $t=4$s & $t=5$s & Avg. \\
\hline

\multicolumn{8}{c}{\textbf{SSIM $\uparrow$}} \\
\hline
Baseline & 0.817 & 0.717 & 0.745 & 0.813 & 0.773 & 0.781 & 0.774 \\
Ours     & 0.878 & 0.830 & 0.818 & 0.873 & 0.867 & 0.898 & \textbf{0.861} \\
\hline

\multicolumn{8}{c}{\textbf{PSNR $\uparrow$}} \\
\hline
Baseline & 25.424 & 20.568 & 22.274 & 24.572 & 22.585 & 22.999 & 23.070 \\
Ours     & 27.994 & 23.423 & 24.933 & 27.754 & 26.967 & 30.358 & \textbf{26.905} \\
\hline

\multicolumn{8}{c}{\textbf{LPIPS $\downarrow$}} \\
\hline
Baseline & 0.075 & 0.184 & 0.188 & 0.135 & 0.158 & 0.168 & 0.151 \\
Ours     & 0.053 & 0.119 & 0.142 & 0.089 & 0.104 & 0.091 & \textbf{0.099} \\
\hline

\end{tabular}
\end{table}

\section{Conclusions}\label{sec:conclusion}
This paper presented SemDynReg, a semantics-guided deformation regularization framework for dynamic 3D Gaussian Splatting. 
The proposed approach leverages semantic cues extracted from SAM and CLIP to establish object-level associations among Gaussians and enforce deformation consistency within the same object. 
By constructing an object-ID map, identifying object-associated top-$k$ Gaussians, and applying object-level regularization to deformation parameters, 
the method introduces structured constraints into otherwise object-agnostic deformation field learning. 
Experimental results on dynamic driving scenes demonstrate that the proposed approach improves deformation consistency of moving objects, 
leading to sharper object appearance, better preservation of fine structures, and enhanced rendering quality compared with the baseline method. 
Future work will investigate extending the proposed regularization framework from object-level modeling to object-part-level modeling, 
enabling more effective handling of non-rigid and articulated objects. 
Such an extension could better capture structured motions in scenarios involving pedestrians, cyclists, robotic manipulators, 
and other objects composed of multiple rigidly moving components.

\bibliographystyle{IEEEtran}
\bibliography{IEEEabrv,Ref}
\end{document}